\documentclass{article} 
\usepackage[preprint]{colm2026_conference}

\usepackage{microtype}
\usepackage{hyperref}
\usepackage{url}
\usepackage{booktabs}
\usepackage{siunitx}
\usepackage{multirow}
\usepackage{graphicx}
\usepackage{tabularx}

\usepackage{lineno}
\usepackage{xspace}
\usepackage{placeins}

\definecolor{darkblue}{rgb}{0, 0, 0.5}
\hypersetup{colorlinks=true, citecolor=darkblue, linkcolor=darkblue, urlcolor=darkblue}

\def\odysseys{Odysseys\xspace}

\title{\odysseys: Benchmarking Web Agents on Realistic \\ Long Horizon Tasks}

\author{Lawrence Keunho Jang\thanks{Equal contribution}\;, Jing Yu Koh\footnotemark[1]\;, Daniel Fried \& Ruslan Salakhutdinov \\
Carnegie Mellon University\\
\texttt{\{ljang,jingyuk,dfried,rsalakhu\}@cs.cmu.edu}
}
\begin{document}
\raggedbottom

\ifcolmsubmission
\linenumbers
\fi

\maketitle

\begin{abstract}
Existing web agent benchmarks have largely converged on short, single-site tasks that frontier models are approaching saturation on. However, real-world web use consists of long-horizon, multi-site workflows. Common web navigation tasks, such as comparing products across different domains, planning trips across multiple services, or summarizing information from multiple search queries, require sustained context and cross-site reasoning over potentially hours of browsing. 
To capture and evaluate such behaviors, we introduce \odysseys: a benchmark of 200 long-horizon web tasks derived from real world browsing sessions evaluated on the live Internet. 
We find that binary pass/fail evaluation is inadequate for long-horizon settings and introduce a rubric-based evaluation, annotating each \odysseys task with an average of 6.1 graded rubrics. We demonstrate that this yields higher agreement with humans and provides a more fine-grained signal than commonly used trajectory-level LLM-as-a-judge evaluation metrics. 
We tested several leading frontier models and find that the strongest models achieve a success rate of 44.5\%, which leaves substantial room for future improvements.
Beyond task success, we argue that efficiency is a first-class concern for long-horizon agents. We introduce a Trajectory Efficiency metric (rubric score per step) and find that even frontier agents achieve only 1.15\%, marking an evident need for agents that can succeed efficiently and not simply eventually.
\odysseys isolates the critical evaluation of long-horizon proficiency in open-web environments, providing a realistic benchmark to measure progress towards computer-use agents that can potentially productively operate for hours. 
We release our tasks, evaluation scripts, and other results at \url{https://odysseys-website.pages.dev}.
\end{abstract}

\section{Introduction}

Recent large language models can now function as computer-use agents. They browse websites, interpret screenshots, click interface elements, and execute multi-step instructions in human-facing software. However, current benchmarks largely evaluate these capabilities in short, tightly scoped episodes, leaving a key regime underexplored: long-horizon web workflows that unfold across many pages, tabs, and domains. This limitation matters because real-world web use is rarely confined to a single page or a single domain. Many common tasks require extended interaction with multiple sites, such as comparing products between retailers, planning travel between booking platforms, or gathering information from search results and synthesizing it into a downstream deliverable. Agents must maintain context over long horizons, reason across heterogeneous websites, decompose open-ended goals into subproblems, and decide when to stop exploring and produce an output.

Towards this goal, we introduce \odysseys, a benchmark of 200 long-horizon web tasks derived from real-world browsing behavior and evaluated on the live Internet. Each task requires agents to execute multi-step workflows across multiple websites. The tasks cover realistic activities and are grounded in annotated human browsing history rather than synthetic templates. 
We also identify the inadequacy of current trajectory-level LLM-as-a-judge evaluation metrics~\citep{he2024webvoyager,xue2025illusion}, commonly used in computer-use benchmarks. 
These benchmarks often rely on LLM judges to evaluate an entire trajectory with a single binary pass-or-fail grade. However, this binary approach becomes increasingly unreliable as tasks grow more complex and action sequences lengthen. To address this, we introduce a rubric-based evaluation system that breaks down each task into a set of specific requirements. Decomposition into rubrics allows the LLM judge to verify individual components more accurately, providing a more informative measure of an agent's partial success. We find that this granular method also aligns more closely with human judgments.

We evaluate the leading frontier and open-weight computer-use agents in \odysseys. The strongest model we tested (Opus 4.6) achieves 44.5\% perfect task success, indicating substantial headroom for progress. Our results and analysis show that current agents struggle with core aspects of long-horizon web navigation. We also observe that several compelling strategies emerge, suggesting that current agents have developed flexible behaviors for operating in open-ended computer environments. 
We conduct an analysis of model performance, test-time scaling with more steps, and do a qualitative analysis and identify common failure modes, highlighting key challenges for future work on building robust long-horizon computer-use agents. Our benchmark is publicly released and maintained at \url{https://odysseys-website.pages.dev}.

\begin{figure}
    \centering
    \includegraphics[width=\linewidth]{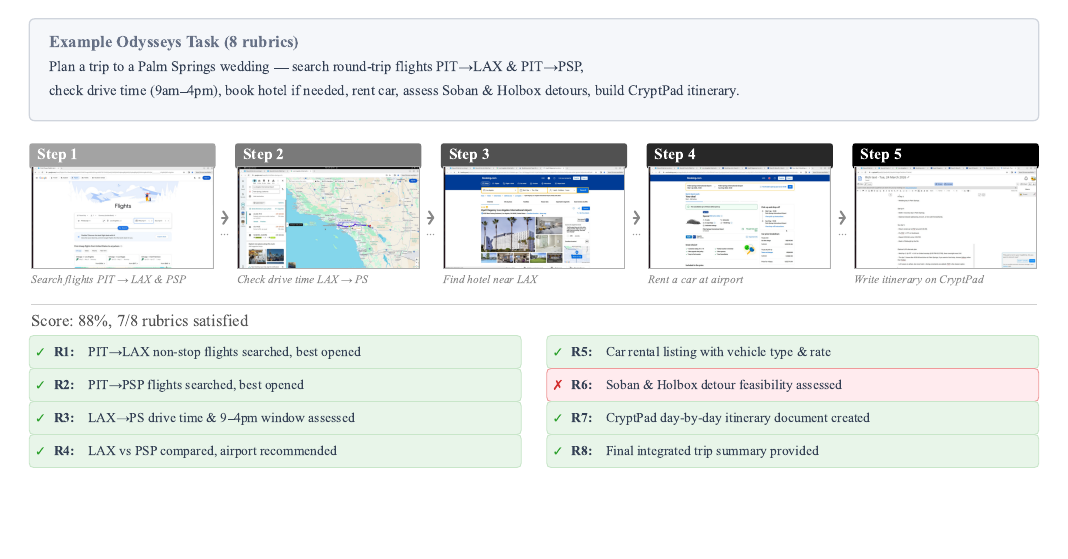}
    \vspace{-0.7in}
    \caption{Odysseys is a long-horizon web agent benchmark with 200 tasks based on real user browsing data. Example task (with simplified task description), trajectory and rubric-based evaluation visualized above.}
    \vspace{-0.1in}
    \label{fig:Fig1}
\end{figure}

\section{Related Work}
\paragraph{Computer-Use Agents}
Benchmarks for computer-use agents based on large language models (LLMs) began with synthetic environments such as MiniWoB~\citep{shi2017world,liu2018reinforcement} and WebShop~\citep{yao2022webshop}. 
More recently, realistic benchmarks such as OSWorld~\citep{xie2024osworld}, Windows Agent Arena~\citep{bonatti2024windows}, Gym-Anything~\citep{aggarwal2026gymanythingturnsoftwareagent} and MacOSWorld~\citep{yang2025macosworld} measure the performance of LLM-based agents on OS-level tasks and office suites.
AndroidWorld~\citep{rawles2024androidworld} and WorkArena~\citep{drouin2024workarena} extend coverage to mobile and enterprise domains respectively, while
AgentBench~\citep{liu2023agentbench} spans web, database, and OS environments.
 
In particular, for web browsing, several benchmarks have been proposed to evaluate GUI-based navigation and task completion, requiring agents to interact through screenshots, clicking, and typing. Mind2Web \citep{deng2023mind2web} introduced the evaluation of LLM agents in web-based tasks using static webpages.
WebArena~\citep{zhou2023webarena} and VisualWebArena~\citep{koh2024visualwebarena} created dynamic synthetic web environments with execution-based verification that is precise but difficult to scale. 
Online-Mind2Web~\citep{xue2025illusion} and WebVoyager~\citep{he2024webvoyager} measure progress with LLM-as-a-judge over full trajectories on real sites on the live internet, which scales but introduces grading noise. Navi-Bench~\citep{yutori2025navigator} alleviates some of these issues by introducing a dynamic task configuration that renders queries and success criteria based on the current date and state of live websites. 
Several static web grounding benchmarks, such as ScreenSpot~\citep{cheng2024seeclickharnessingguigrounding, li2025screenspotproguigroundingprofessional}, OmniACT~\citep{kapoor2024omniactdatasetbenchmarkenabling}, and OSWorld-G~\citep{xie2025scalingcomputerusegroundinguser}, also provided evaluation measures for visual grounding specific metrics.
At longer research horizons, GAIA~\citep{mialon2023gaiabenchmarkgeneralai} and BrowseComp~\citep{wei2025browsecompsimplechallengingbenchmark} evaluate multi-hop information synthesis from the open web, targeting the deep research paradigm, but shift away from GUI-level interaction toward API-based or text-based access, evaluating final answers rather than the browsing process itself.
Critically, all of these web benchmarks target short-answer or single-site tasks (which are saturated by frontier models) or do not truly test the full multimodal and navigation capabilities of a computer-use agent.
 
\paragraph{Long-Horizon LLM Agents}
As LLMs have increased their general capabilities to reason and act as autonomous agents, their ability to expand into long-horizon tasks has become a central focus. METR~\citep{kwa2026measuringaiabilitycomplete} introduces a task-completion time horizon as a measure of long-horizon capability, showing that the effective task duration of frontier agents (measured by human completion time) has grown exponentially from 2019 to 2025. Despite this progress, current agents remain limited to tasks on the order of hours and still fall short of reliably completing longer, multi-hour or day-scale workflows. 
Early work on multi-step reasoning of LLMs, such as HotpotQA \citep{yang2018hotpotqadatasetdiverseexplainable} and MuSiQue \citep{trivedi2022musiquemultihopquestionssinglehop}, studied multi-hop reasoning over multiple documents but mainly operated in text, static settings without sequential decision-making or environment interaction. 
Several long-horizon agentic benchmarks have also been released:
$\tau$-bench~\citep{yao2024taubenchbenchmarktoolagentuserinteraction} measures consistency in multi-turn tool calling agents;
Vending-Bench~\citep{backlund2025vendingbenchbenchmarklongtermcoherence} tests LLM agents on their ability to run a vending machine over a long time period;
TheAgentCompany~\citep{xu2025theagentcompanybenchmarkingllmagents} evaluates multi-platform workplace tasks in a simulated software company;
and TravelPlanner~\citep{xie2024travelplannerbenchmarkrealworldplanning} evaluates the collapse of LLMs in large-scale multi-constraint planning.
OdysseyBench~\citep{wang2025odysseybenchevaluatingllmagents} extends this line to long-horizon office workflows across multiple applications. 

\section{The Odysseys Dataset}

\begin{table*}[t]
\centering
\small
\begin{tabularx}{\textwidth}{@{}cXX@{}}
\toprule
\textbf{Level} & \textbf{Task Summary} & \textbf{Example Rubric Criteria} \\
\midrule

\raisebox{-4.7\normalbaselineskip}[0pt][0pt]{\rotatebox[origin=c]{90}{\textbf{Easy}}} &
Find a roasted Brussels sprouts recipe with Parmesan and balsamic vinegar; check Le Creuset for a 5.5\,qt Dutch oven in light green; find a YouTube video on dishwasher air gap installation. &
-- Recipe page shows both Parmesan and balsamic vinegar with title, oven temp, and cook time \newline
-- Le Creuset product page confirms whether 5.5\,qt is available in light green \\
\cmidrule{2-3}

&
Build a weather risk snapshot: get Baltimore current conditions, Syracuse 7-day low from Wunderground, NWS forecast and alerts for Rittman, OH, and predicted snowfall from NWS Mount Holly. &
-- Report the lowest forecasted temperature in Syracuse over the next 7 days from Wunderground \newline
-- Report predicted snowfall from the NWS Mount Holly winter forecast graphic \\

\midrule

\raisebox{-4.8\normalbaselineskip}[0pt][0pt]{\rotatebox[origin=c]{90}{\textbf{Medium}}} &
Plan a trip to Japan as a Canadian in the U.S.: find the passport renewal process, fees, and processing times; check visa requirements; review travel advisories; compare travel insurance from PolicyAdvisor and Kanetix. &
-- Correctly identify the official Canadian passport renewal process for citizens abroad \newline
-- Determine whether a Canadian passport holder needs a tourist visa for Japan \\
\cmidrule{2-3}

&
Evaluate a Tesla Model 3 lease in Los Angeles: find current lease terms and price, search Craigslist for trailer alternatives less than \$10K, find a comparable Zillow rental, and present a cost comparison. &
-- Summarize current LA Tesla Model 3 lease offer with payment, term, and due-at-signing \newline
-- Identify $\geq$3 Craigslist trailer listings under \$10K suitable for living \\

\midrule

\raisebox{-4.7\normalbaselineskip}[0pt][0pt]{\rotatebox[origin=c]{90}{\textbf{Hard}}} &
Plan a trip to all 30 MLB stadiums: select one real game date per stadium from the official schedule, prioritize star players, plan travel logistics, find accommodations and local food via Yelp, and compile in CryptPad. &
-- Itinerary includes all 30 stadiums with one official game date and matchup each \newline
-- Complete visit sequence with cheaper travel mode identified between all stops \\
\cmidrule{2-3}

&
Find the top 10 banana bread recipes by popularity, compare ingredients and methods, select 3 most unique, find associated YouTube videos, and synthesize a combined best recipe explaining design choices. &
-- Ingredients and cooking methods extracted and compared across all 10 recipes \newline
-- Combined recipe created drawing from strengths of top 3, with complete instructions \\

\bottomrule
\end{tabularx}
\vspace{-0.1in}
\caption{Representative tasks from \odysseys spanning three difficulty levels and diverse real-world domains. Task descriptions are condensed for readability to capture key actions (full descriptions provided in Appendix~\ref{app:examples}). Each task requires the agent to navigate multiple websites, extract specific information, and satisfy verifiable rubric criteria.}
\vspace{-0.1in}
\label{tab:odyssey-examples}
\end{table*}

We introduce the \odysseys dataset, consisting of 200 long-horizon, multi-site web tasks designed to evaluate web agents on realistic browsing workflows. Each task begins from a Google search page and requires the agent to navigate across multiple websites to complete complex, multi-step objectives such as comparing products across e-commerce platforms, planning travel itineraries, or setting up video playlists and watching lectures. We show an overview of a few example tasks in Table~\ref{tab:odyssey-examples}.

\subsection{Collection Process}

We recruit 248 participants based in the US and UK through the Prolific (\url{https://prolific.com}) crowdsourcing platform and provide them with a desktop application that reads their Chrome browsing history and provides the user with an interface to annotate their own history into web agent tasks. The application segments browsing activity into singular tasks with corresponding URLs using Chrome's Journey algorithm \citep{google2022chrome_journeys}. The collection yields 2,380 labeled journeys that span various domains, from comparison shopping and travel planning to media consumption and information research. A summary and visualization of our annotation interface is provided in Appendix~\ref{app:journeys-ui}.

\subsection{From Journeys to Odysseys} \label{sec:task-composition}

Raw journey labels are noisy, as participants may submit labels unrelated to the URLs, underspecify tasks, or provide vague descriptions. We address this through a two-stage refinement process combining LLM screening with manual review by the authors (see Appendix~\ref{app:refinement} for details). After filtering for label accuracy, feasibility, login requirements, and overall quality, 696 high quality journeys (29.2\%) remained.

The resulting journeys represent real browsing behavior but are generally short tasks due to Chrome's journey segmentation. To compose them into realistic long-horizon tasks, we cluster related journeys by embedding similarity and use GPT 5.4 to arrange subsets into coherent multi-step workflows (see Appendix~\ref{app:composition-pipeline} for full details of the clustering, composition pipeline, and prompt). For each composed task, the LLM generates a natural-language prompt, step plan, rubric with verification procedures, and a self-assessed coherence score. Each source journey is used at most 3 times across all \odysseys to prevent over-representation, and the authors manually perform quality assurance by inspecting every long chained Odyssey and de-duplicating any tasks that appear too similar (see Fig.~\ref{fig:odysseys-qa}).
We reject generated tasks that span fewer than 2 sites, have incomplete rubrics, or score below 2/5 on coherence. Difficulty is assigned by step count and domain spread: \textit{easy} ($\leq$5 steps, $\leq$3 domains), \textit{medium} (6--8 steps or 4+ domains), \textit{hard} (exceeding both).
In addition to the 90 composed tasks, the authors of this paper contributed 30 manually authored long-horizon tasks grounded in real personal queries, following the same schema for a total of 200 tasks. We also used an LLM pipeline to generate an additional 80 hard Odysseys tasks by providing GPT 5.4 with in-context examples of human-authored long-horizon tasks, which we got human annotators to review and filter down (see Appendix~\ref{app:hard-longhorizon-cua-pipeline}).
\subsection{Rubrics}
\label{sec:rubrics}

Prior work on computer-use agents typically leverages execution-based verification~\citep{zhou2023webarena,koh2024visualwebarena,xie2024osworld} which requires handcrafted task-specific rewards, or LLM-as-a-judge over web agent trajectories~\citep{he2024webvoyager,xue2025illusion} where an LLM takes in the task instruction, screenshots and actions of an agent execution trace, and outputs a score or a pass/fail metric. However, we find that these are not suitable for \odysseys, due to the long-term nature of the tasks, as well as the difficulty of crafting rules-based rewards for each task. 
Inspired by prior work on rubrics and checklists for evaluating and training language models~\citep{shao2025dr,hashemi2024llm,gunjal2025rubrics,viswanathan2025checklists, aggarwal2026gymanythingturnsoftwareagent}, we investigate using rubric-based evaluation for \odysseys tasks. 

For each task, we generate a structured set of rubrics to evaluate partial progress (examples in Fig.~\ref{fig:Fig1}, Fig.~\ref{fig:gpt-14-pastebase64} and Table~\ref{tab:odyssey-examples}). 
Rubrics are generated alongside the task during LLM composition. 
For each rubric item, the model produces a \textit{requirement} (a single verifiable checkpoint), and  a \textit{verification} description of how a grader should determine whether it is accomplished.
Tasks contain 3--12 rubric items, with an average of 6.1 rubrics per task. All rubric items are verified by an author for correctness and relevance to task completion, and we also run an LLM with a text web search tool to cross-check verification criteria against live websites.

To evaluate model performance with rubrics, we prompt an LLM-as-a-judge given each all screenshots and actions in the trajectory, as well as the rubric description. We use \texttt{gemini-3.1-flash-lite-preview} as the judge for all our experiments, and provide the full prompts used in Appendix~\ref{app:rubric-judge-prompt}. Each rubric is evaluated separately using the judge. For each task, we report the averaged and perfect (binary, 1 if all rubrics are 1) rubric scores.

\subsection{Final Review}
After chaining, all task prompts are \textit{CUAfied}: every task is rewritten from abstract 
information-gathering descriptions into natural, conversational requests styled as messages 
to a computer-use agent. Additionally, the LLM rewrites each prompt to include specific sites with concrete parameters (e.g., zip codes, price ranges, filter criteria), first-person motivation, natural 
step chaining without numbered lists, and visual browser-proof requirements such as keeping 
pages open in tabs, adding items to carts, or opening side-by-side comparisons. The rubric, 
steps, dependencies, and deliverable are regenerated in the same pass to align with the 
rewritten prompt. This ensures that  each task reads as a realistic user request and that the evaluation criteria reference observable 
browser states. 

Every task, composed or manually authored, undergoes a two-pass review using a dedicated QA interface (Fig.~\ref{fig:odysseys-qa}). In the first pass, the authors verify prompt coherence, trace each component to its source journey, and ensure that rubric items are unambiguous and actionable. All prompts that reveal any potential personally identifiable information (PII) are also rewritten or removed. In the second pass, an LLM-augmented pass uses a web search tool to confirm site accessibility and step feasibility; a human adjudicates flagged issues. Tasks that fail in either pass are revised or removed. 
After post-processing, the \odysseys benchmark contains 200 tasks.

\subsection{Dataset Statistics}

The benchmark comprises 200 tasks spanning three difficulty levels: easy (45), medium (46), and hard (109). Task instructions average 272.3 words (median 277.5, range 76–387), reflecting the detailed, multi-step nature of real-world web workflows. The tasks are evaluated using a total of 1,225 rubric items, with an average of 6.1 rubrics per task (median 6, range 3–12). The data set spans 22 top-level domains and 88 fine-grained SimilarWeb categories, with each task touching an average of 1.9 categories (up to 3). The most represented domains are Ecommerce and Shopping (43 tasks), Travel and Tourism (38), Science and Education (37), and Computers, Electronics and Technology (34), followed by Lifestyle (21), Health (21), Community and Society (20), Arts and Entertainment (19), and Food and Drink (18).

\section{Experiments}

We benchmark several frontier API-based models and open weight models on \odysseys. Each model is implemented with their own recommended settings for computer-use, and we launch them in the same virtual Ubuntu environments from OSWorld~\citep{xie2024osworld}. The primary application that the models use is Google Chrome, although some models also occasionally choose to use other applications (such as LibreOffice for generating a report). 
We only benchmarked models that have general computer-use abilities, and do not test models that are only trained for web navigation, as many \odysseys tasks involve subtasks that are infeasible for web navigation only models (e.g., maintaining multiple tabs).

\begin{table}[t]
\centering
\resizebox{\textwidth}{!}{%
\begin{tabular}{@{}ll
                S[table-format=2.1]
                S[table-format=2.1]
                S[table-format=2.1]
                S[table-format=2.1]
                S[table-format=1.2]@{}}
\toprule
\multirow{2}{*}[-2pt]{\textbf{Type}} & \multirow{2}{*}[-2pt]{\textbf{Model}}
& \multicolumn{1}{c}{\multirow{1}{*}[-4pt]{\textbf{O-M2W Judge}}}
& \multicolumn{2}{c}{\textbf{Rubrics (\%) ($\uparrow$)}}
& \multicolumn{1}{c}{\multirow{1}{*}[-4pt]{\textbf{\# Steps}}}
& \multicolumn{1}{c}{\multirow{1}{*}[-4pt]{\textbf{Traj.\ Eff.}}} \\
\cmidrule(lr){4-5}
& & \textbf{($\uparrow$)} & \multicolumn{1}{c}{Averaged} & \multicolumn{1}{c}{Perfect} &  \textbf{($\downarrow$)} & \textbf{(\%)\ ($\uparrow$)} \\
\midrule
\multirow{4}{*}{API}
& Opus 4.6     & \textbf{37.0} & \textbf{68.9} & \textbf{44.5} & 81.3 & 1.06 \\
& GPT-5.4      & 21.5 & 55.4 & 33.5 & 64.4 & \textbf{1.15}\ \\
& Sonnet 4.6   & 25.0 & 49.8 & 31.0 & 80.4 & 0.79 \\
& GPT-5.4-mini & 15.5 & 38.4 & 10.5 & \textbf{41.7} & 1.12 \\
\midrule
\multirow{4}{*}{Open Weights}
& Qwen-3.5-9B                    & 18.0 & 42.6 & 13.5 & 78.3 & 0.75 \\
& Qwen-3.5-4B                    & 18.3 & 42.9 & 10.7 & 86.4 & 0.69 \\
& Qwen-3.5-35B-A3B & 9.5 & 28.5 & 6.5 & 86.1 & 0.42 \\
& UI-TARS-1.5-7B                & 2.0 & 10.0 & 1.0 & 76.6 & 0.23 \\
\bottomrule
\end{tabular}%
}
\vspace{-0.12in}
\caption{Model performance on \odysseys.}
\vspace{-0.05in}
\label{tab:model-scores}
\end{table}

\subsection{Evaluation} \label{sec:eval}

\begin{table}[t]
\centering
\resizebox{\textwidth}{!}{%
\begin{tabular}{lcccccc}
\toprule
\textbf{Metric}  & \textbf{Cohen's $\kappa$ ($\uparrow$)} & \textbf{Precision ($\uparrow$)}  & \textbf{Recall ($\uparrow$)} & \textbf{F1 ($\uparrow$)} & \textbf{Accuracy ($\uparrow$)} \\
\midrule
O-M2W Judge & 0.508  & 0.842  & 0.696  & 0.762 & 0.752 \\
\midrule
Rubrics (Averaged) \hspace{8mm}      & 0.788 & \textbf{0.939}  & 0.960 & \textbf{0.949} & 0.923 \\
Rubrics (Perfect)      & \textbf{0.849} & 0.901  & \textbf{0.970}  & 0.934 & \textbf{0.926} \\
\bottomrule
\end{tabular}
}
\vspace{-0.1in}
\caption{Automated LLM-based metrics vs.\ human agreement. Rubrics scores (Cohen's $\kappa$, precision, recall, F1, and accuracy against human labels) show substantial agreement, while the trajectory-level Online-M2W (holistic binary judgment) agreement is worse.} 
\vspace{-0.1in}
\label{tab:judge-correlation}
\end{table}

\begin{figure}[t]
    \centering
    \includegraphics[width=\linewidth]{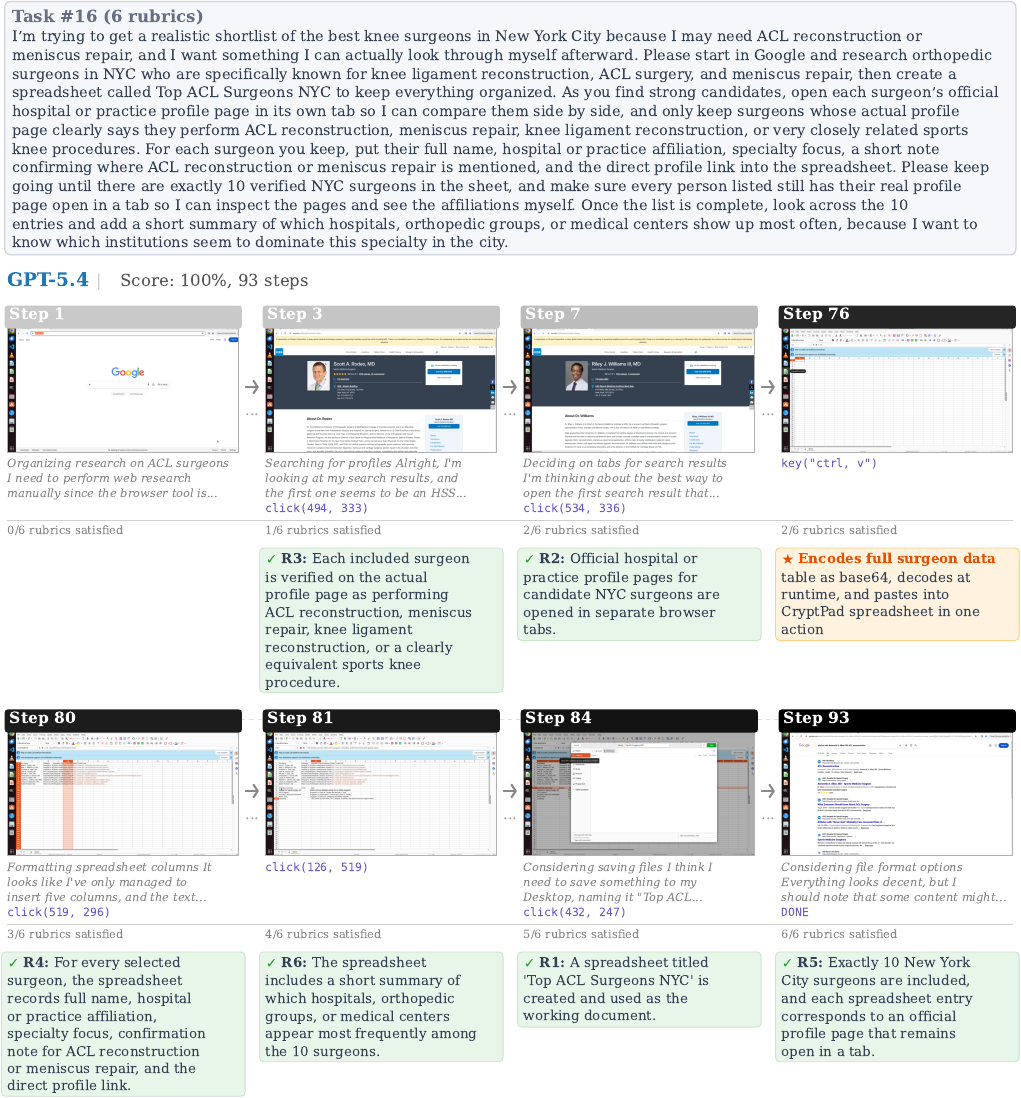}
    \vspace{-0.25in}
    \caption{After researching 10 surgeon profiles, GPT-5.4 encoded the entire data table as a base64 string inside the Python action, decoded it at runtime with \texttt{pyperclip.copy(\_text); pyautogui.hotkey('ctrl', 'v')}. This preserves exact tab and newline characters, and populates the spreadsheet in a single paste with the correct column mapping.}
    \vspace{-0.1in}
    \label{fig:gpt-14-pastebase64}
\end{figure}

\paragraph{Rubrics Scores} As discussed in Sect.~\ref{sec:rubrics}, we compute rubric scores based on the specific success criteria of each task. To determine the effectiveness of this metric, we measure the agreement between automated LLM judges and human annotations at three levels of granularity, summarized in Table~\ref{tab:judge-correlation}. At the finest level, \textit{rubrics (average)} treats each task--rubric pair as an independent observation and computes agreement across all 1,225 rubric pairs; for the \textit{rubrics (perfect)} metric, a task is scored as passing only if every one of its rubrics are satisfied, yielding a single binary label per task; and at the \textit{trajectory} level, we run the LLM judge from Online-M2W~\citep{xue2025illusion}, which issues a holistic pass/fail judgment based on the agent's action trajectory.
We collected human annotations of a subset of 120 Opus 4.6 trajectories, where each annotator rated whether each rubric was satisfied during the course of the trajectory (see Appendix~\ref{app:human-agreement-ui} for a visualization of the annotation interface). Then, we compute human agreement of each metric by measuring Cohen's $\kappa$ coefficient, the F1 score, and the accuracy against the collected human labels.
Across all metrics, the rubric-based evaluations substantially outperform the trajectory judge: rubric-average agreement achieves a Cohen's $\kappa$ of 0.788 and F1 of 0.949, compared to 0.508 and 0.762 for the trajectory judge. The task-level \emph{perfect} metric shows even higher agreement ($\kappa = 0.849$) but slightly lower $F1 = 0.934$. 
Additionally, rubrics also provide partial credit for weaker models, which may only achieve zero scores under the trajectory-level judge. 
This also introduces a meaningful reward signal for training weaker models with reinforcement learning, which would be a strong direction for future work.

\paragraph{Trajectory Efficiency.}
Raw rubric scores capture \emph{what} an agent accomplished but not \emph{how efficiently} it performs. This matters for \odysseys in particular, as a 100-step trajectory takes Opus 4.6 roughly 30 minutes of wall-clock time to complete, which is far longer than a human would realistically wait for any of these tasks to finish. Two agents may reach the same rubric score, but one that does so in 30 steps delivers a dramatically better user experience than one that does so in 100. To quantify this, we report \emph{Trajectory Efficiency}, defined as the per-task ratio of rubric score to step count, averaged over all tasks:
\begin{equation}
  \text{Traj.\ Eff.} = \frac{1}{N}\sum_{i=1}^{N} \frac{s_i}{n_i},
\end{equation}
where $s_i$ is the averaged rubric score for task~$i$ and $n_i$ is the number of agent steps taken on that task. Unlike the Success weighted by Path Length metric from Vision-Language Navigation~\citep{anderson2018evaluation}, we do not know the oracle number of steps required for each task and therefore cannot normalize by an optimal trajectory length. Higher Trajectory Efficiency indicates that an agent achieves strong outcomes in fewer steps, penalizing trajectories that eventually succeed only after substantial wasted effort.

Step count in the denominator above counts environment steps. Step counts is the number of LLM calls, once for each observation-action loop step within the web agent runner. GPT-5.4 averages 2.03 actions per call (49.1\% of calls return two or more actions, up to 24 in a single turn), whereas Opus 4.6 returns exactly 1 action per call in every task. GPT-5.4 therefore spends roughly half as many model round-trips, which materially reduces wall-clock time per task. Qualitatively, GPT-5.4's batched turns often forecast the next few UI states, such as typing a query and pressing enter, and queue the corresponding actions ahead of a fresh screenshot. Opus's per-call observe/act loop forgoes this shortcut even when the next action is trivial. This gap suggests that training or prompting for multi-action batching is a practical path for efficiency on long-horizon tasks.

\subsection{Results}
\begin{figure}
    \centering
    \includegraphics[width=0.95\linewidth]{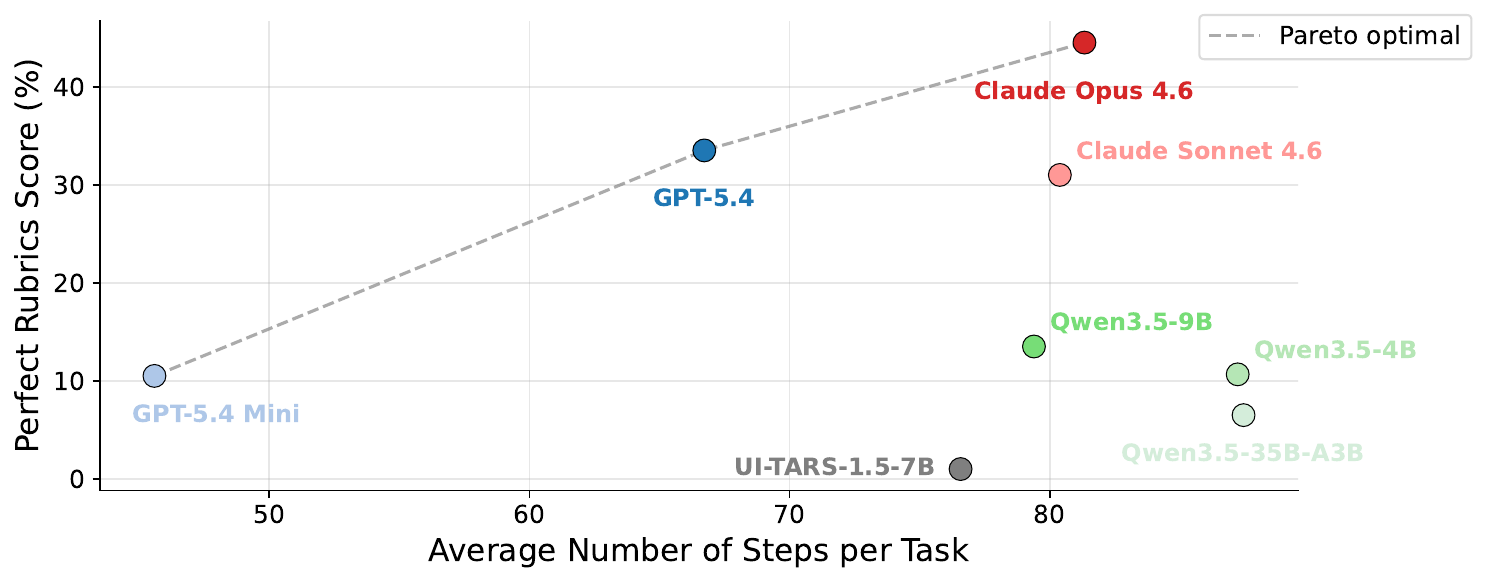}
    \vspace{-0.15in}
    \caption{A scatter plot of perfect success rate against number of steps taken. The pareto optimality curve for models we tested is also shown.}
    \vspace{-0.1in}
    \label{fig:success-steps-pareto}
\end{figure}

We run all experiments under the same settings: we limit the number of steps to at most 100 steps\footnote{For 100 steps, Opus 4.6 takes approximately 30 mins and \$2.5 to run per task.}, and we use the max reasoning effort (if any). We use the runner from OSWorld~\citep{xie2024osworld}, and start the models off with a Google Chrome window at the provided starting url for each task (if any). The models have access to the full Google Chrome interface, and are allowed to open tabs, save files, or do anything else supported by the virtual machine. 
We evaluate all models with the proposed rubric-based evaluation (Sec.~\ref{sec:rubrics}). The results are summarized in Table~\ref{tab:model-scores}. We observe that the best frontier models trained with computer use capabilities (Opus 4.6, GPT-5.4) achieve the strongest scores, with Opus 4.6 scoring slightly better across most metrics. We observed that GPT-5.4 models generally use fewer steps, which can be more Pareto optimal (see Fig.~\ref{fig:success-steps-pareto}). 
Open weight models trail significantly behind frontier models, with the Qwen-3.5-9B model performing the best amongst the open weights CUAs that we evaluated.

\FloatBarrier
\subsection{Analysis} \label{sec:analysis}


\begin{figure}[t]
    \centering
    \includegraphics[width=\linewidth]{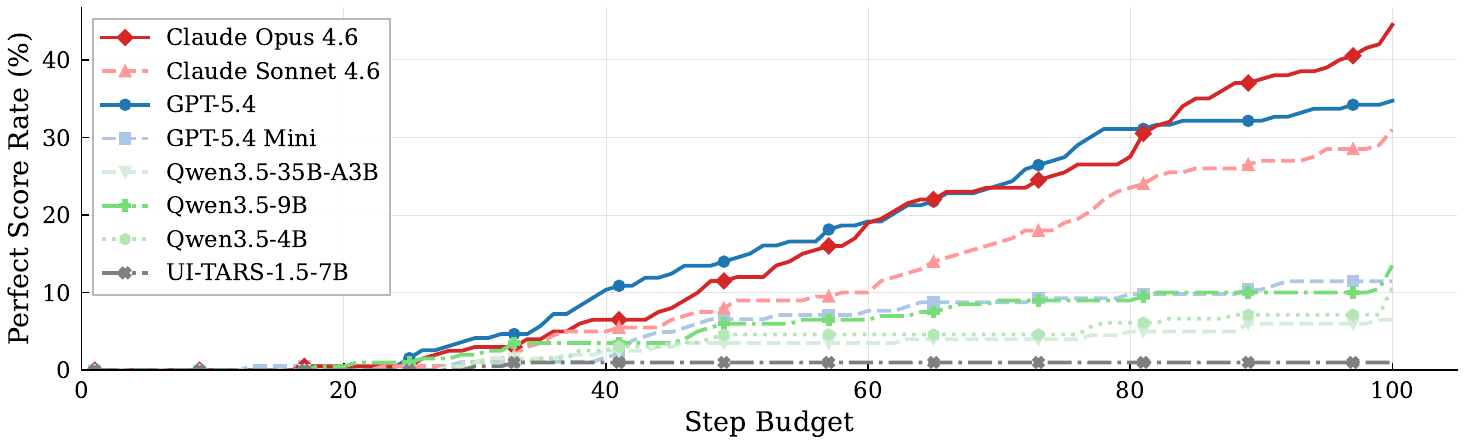}
    \vspace{-0.3in}
    \caption{Perfect rubric rate as a function of step budget. Each curve shows the mean rubrics score achieved within a given number of interaction steps.}
    \vspace{-0.1in}
    \label{fig:success-vs-steps}
\end{figure}

\begin{figure}[t]
    \centering
    \includegraphics[width=\linewidth]{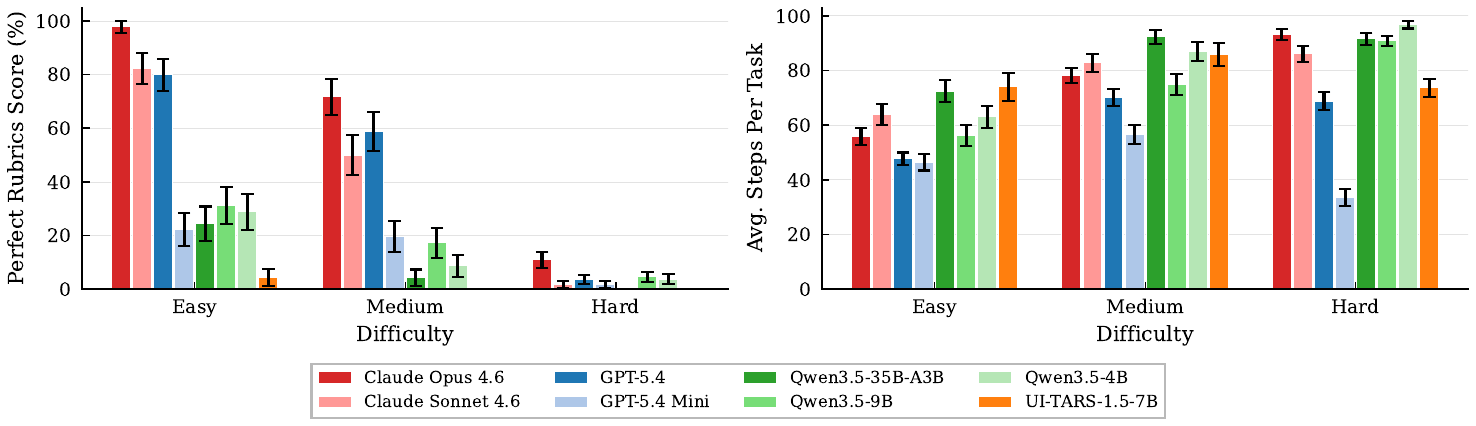}
    \vspace{-0.3in}
    \caption{Rubric scores (left) and \# of steps taken per task (right) broken down by difficulty.}
    \vspace{-0.1in}
    \label{fig:difficulty-bars}
\end{figure}

\begin{table}[t]
\centering
\small
\setlength{\tabcolsep}{6pt}
\begin{tabular}{@{}l
S[table-format=3.1]
S[table-format=3.1]
S[table-format=3.1]
S[table-format=3.1]
@{}}
\toprule
\multirow{2}{*}{\textbf{Model}} &
\multicolumn{1}{c}{\textbf{Easy}} &
\multicolumn{1}{c}{\textbf{Medium}} &
\multicolumn{1}{c}{\textbf{Hard}} &
\multicolumn{1}{c}{\textbf{Overall}} \\
&
\multicolumn{1}{c}{(n=45)} &
\multicolumn{1}{c}{(n=46)} &
\multicolumn{1}{c}{(n=109)} &
\multicolumn{1}{c}{(n=200)} \\
\midrule
Claude Opus 4.6 & 97.8 & 71.7 & 11.0 & 44.5 \\
Claude Sonnet 4.6 & 82.2 & 50.0 & 1.8 & 31.0 \\
GPT-5.4 & 80.0 & 58.7 & 3.7 & 33.5 \\
GPT-5.4 Mini & 22.2 & 19.6 & 1.8 & 10.5 \\
Qwen3.5-35B-A3B & 24.4 & 4.3 & 0.0 & 6.5 \\
Qwen3.5-9B & 31.1 & 17.4 & 4.6 & 13.5 \\
Qwen3.5-4B & 28.9 & 8.7 & 3.8 & 10.7 \\
UI-TARS-1.5-7B & 4.4 & 0.0 & 0.0 & 1.0 \\
\bottomrule
\end{tabular}
\vspace{-0.1in}
\caption{Perfect rubrics rate (\%) by difficulty on the Odysseys benchmark.}
\vspace{-0.1in}
\label{tab:results-by-difficulty}
\end{table}

\paragraph{Difficulty levels} As described in Sec.~\ref{sec:task-composition}, we also categorized \odysseys tasks by their difficulty levels, which are roughly correlated with how many steps are required to solve the task, and how many unique website domains it needs to traverse. We summarize the performance of the models on these tasks in Fig.~\ref{fig:difficulty-bars} and Table~\ref{tab:results-by-difficulty}. 
We observe that performance substantially degrades on hard tasks, with Opus 4.6 averaging a rubric score average of 53.2 and a perfect rate of 11.0\%, substantially outperforming GPT-5.4 (which achieves a score of 33.8) and a perfect rate of 3.7\%. Opus 4.6 and Sonnet 4.6 models also tend to use more steps across all difficulty levels, particularly on hard tasks. 

\paragraph{Scaling with step budget} Fig.~\ref{fig:success-vs-steps} shows how task completion improves as models are allowed more interaction steps. All models exhibit a broadly sigmoidal scaling curve, as the perfect score rates remain close to zero for the first ${\sim}15$ steps, reflecting the minimum number of interactions needed to complete the simplest \odysseys tasks. Rates then rise steadily through the 20--70 step range as progressively harder tasks are fully solved, before the rate of improvement tapers past ${\sim}80$ steps. The two frontier models, Claude Opus 4.6 and GPT-5.4, reach substantially higher asymptotic perfect score rates and climb more steeply through the mid-range, indicating greater per-step efficiency in addition to a higher capability ceiling. In contrast, GPT-5.4 Mini and Qwen 3.5 (35B-A3B and 9B) exhibit markedly shallower slopes and lower asymptotes, suggesting their shortfall reflects fundamental capability gaps rather than insufficient step budgets. We perform an ablation on increasing the max steps to  200, where we find Claude Opus 4.6 to eventually complete tasks at a much higher rate, but Qwen 3.5 is not able to (See \ref{app:200step-scaling}). This finding is promising for the capability of current web agents, but shows room for improvement in trajectory efficiency.

\paragraph{Unique failure modes per model}
All frontier models we tested exhibit distinct failure patterns on long-horizon tasks. Opus 4.6 most commonly fails through over-investment in research. It continues to gather information, but does not transition to producing the required deliverable, such as creating the final document, and eventually exhausts the step budget with an empty output artifact. This pattern appears in 6 of its 12 zero-score runs. This results in Opus runs remaining productive but incomplete until termination, hitting the 100-step cap on 39\% of tasks, while its partial-credit runs average 97.4 steps.

GPT-5.4, by contrast, more often fails through inaction despite correct high-level reasoning. On 4 of its 7 zero-score tasks, it generates long and detailed plans that correctly identify which pages to visit and what information to collect, but then terminates with an empty action after only a few steps, sometimes without any browser interaction at all. 
The two models also share a common failure mode on especially broad tasks with many parallel subtasks. Tasks that require visiting 10 to 30 venues, such as planning a trip to every MLB stadium, often cause both models to become stuck in the first phase of the task, such as collecting schedules, without progressing to later stages like flight search, hotel selection, or document compilation. All such tasks receive zero scores from both models. This suggests that current agents struggle not only with long sequential horizons, but also with high-fanout task structure, where they must allocate effort across many related subtasks. This is one direction for which models would likely benefit from subagents or a multi-agent setup.

\paragraph{Surprising capabilities}
Both GPT-5.4 and Opus 4.6 models exhibit behaviors that were notably sophisticated. GPT-5.4 in particular develops several strategies that repurpose browser and system primitives for information extraction. We observed it using an unprompted strategy for bulk spreadsheet entry: rather than pasting cell contents directly, it encodes full tables as base64 strings inside its Python action, decodes them at runtime, and pastes the result through the clipboard so that tab and newline delimiters are preserved (Fig.~\ref{fig:gpt-14-pastebase64}). We observe this in 8 runs. Another recurring pattern is that when a product page fails to render correctly due to JavaScript errors, the model navigates to the raw HTML through the \texttt{view-source:} protocol, searches within the source using \texttt{ctrl+f}, and extracts structured product metadata from embedded JSON-LD markup, including variant identifiers, stock status, and size mappings (Fig.~\ref{fig:gpt-62-viewsource}). We observe this behavior in 23 runs. 

Opus 4.6 exhibits a different set of strengths. When target websites return 403 errors, it sometimes falls back to the Wayback Machine by constructing archived URLs and retrieving cached versions of otherwise inaccessible pages (Fig.~\ref{fig:claude-48_wayback}). This allows it, for example, to recover class schedule information that would not be available through direct browsing alone. Opus also makes substantially heavier use of middle-click to open links in background tabs without losing its place on the current page, using this interaction 131 times across all runs compared to just 7 for GPT-5.4. In addition, it uses \texttt{ctrl+f} not only to locate information, but also to falsify hypotheses about page relevance. After searching for a keyword and observing no matches, it often treats that absence as evidence that the page is unproductive and moves on immediately (Fig.~\ref{fig:claude-29_ctrlf_probe}).

\section{Conclusion}

In this work we introduced \odysseys, a benchmark for evaluating web agents on realistic long-horizon tasks drawn from real browsing behavior. In contrast to prior benchmarks that emphasize short, single-site episodes, \odysseys focuses on long multi-site workflows that require sustained planning, navigation, and cross-page reasoning. 
We demonstrated that existing trajectory-level evaluation methods were insufficient for long-horizon trajectories, and proposed rubric-based evaluations which decompose long-horizon success into verifiable intermediate outcomes. This yields substantially higher agreement with human judgment. 
Our experiments show that, despite strong recent progress in computer-use agents, long-horizon web interaction is still unsolved. Even the best frontier models achieve only about 44.5\% perfect-task success, with performance dropping sharply on harder tasks and plateauing as step budgets increase.
Our analysis suggests that current limitations reflect deeper challenges in long-context planning, maintaining coherence across sites, and reliably executing extended workflows. Improving computer-use agents on long-horizon tasks, such as through reinforcement learning, inference-time search, or the efficiency-oriented techniques above, is a promising direction for future work. We release \odysseys tasks, rubrics, and analysis publicly at \url{https://odysseys-website.pages.dev}. We believe the combination of realistic task design and fine-grained evaluation in \odysseys will help drive future progress on computer-use agents that are able to operate robustly over extended time horizons.

\pagebreak
\section*{Acknowledgments}
We would like to thank Geronimo Carom, Andrew Jang, and Mareks Woodside for participating in human agreement annotation studies. We also thank Wendy Kua for contributing several tasks. We thank Naveen Raman for inspiring conversation in the office and Maxwell Jones for submitting journeys while curating vegetable-free lab lunches throughout the semester.

\section*{Ethics Statement}

\paragraph{Intended uses} \odysseys is a research benchmark designed to measure and evaluate the progress of computer-use agents on long-horizon web tasks. The models and scaffolds evaluated in this paper are research prototypes assessed in controlled environments, and our results are not intended to endorse their deployment in practical applications at present. Although tasks in \odysseys are executed on the live Internet for the best measure of real world behavior of computer-use agents, the tasks are designed around read-only or low-impact interactions such as browsing, searching, and comparing information, rather than actions with real-world side effects such as completing purchases or submitting forms.

\paragraph{Potential for misuse} As computer-use agents become more capable, they could be leveraged for malicious purposes such as automated scraping of personal information, more sophisticated phishing, or circumventing website access controls. We note several emergent behaviors in our analysis in Sec.~\ref{sec:analysis}, such as using \texttt{view-source:} to extract structured metadata from pages that fail to render, or retrieving cached content via the Wayback Machine to bypass access restrictions. These illustrate that agents can repurpose browser primitives in unintended ways. Developers deploying such agents should consider these capabilities and implement appropriate safeguards, including respecting website terms of service and robots.txt policies.

\paragraph{Broader impacts} Advancing autonomous web agents has the potential to improve accessibility for users with disabilities or limited technical skills, and to automate tedious, repetitive computer workflows. However, as our results show that even frontier models achieve only 44.5\% perfect task success on \odysseys, and exhibit several key failure modes such as those described above. As their capabilities improve, researchers and developers should carefully consider the economic and social implications, including potential effects on employment. We believe that benchmarks like \odysseys, which expose concrete capability gaps in frontier models, contribute to the responsible development of this technology by enabling the community to measure progress transparently.

\paragraph{Data collection and privacy} The browsing history used to construct \odysseys was collected from consenting participants recruited through Prolific, who voluntarily annotated their own Chrome browsing histories. All task prompts were reviewed and any content that revealed personally identifiable information (PII) was rewritten or removed. No raw browsing histories are released; only the final, de-identified task prompts and rubrics are published.

\bibliography{colm2026_conference}
\bibliographystyle{colm2026_conference}

\pagebreak
\appendix
\section{Appendix}

\subsection{Large Language Model Disclosure}

We iteratively use a coding agent (Claude Code) with a human in the loop to generate several plots in this paper conditioned on numerical results and visualization instructions. We also used Claude Code to flag potentially interesting examples in the trajectories and write a description of these examples. We reviewed these outputs and polished these for our qualitative analysis section of the paper. We also give Claude Code access to our codebase, results, and data, to flag any issues with our paper and provide feedback for iterative writing. 


\subsection{Data Collection Interface} \label{app:annotation-ui} \label{app:journeys-ui}

\begin{figure}[h]
    \centering
    \includegraphics[width=\linewidth]{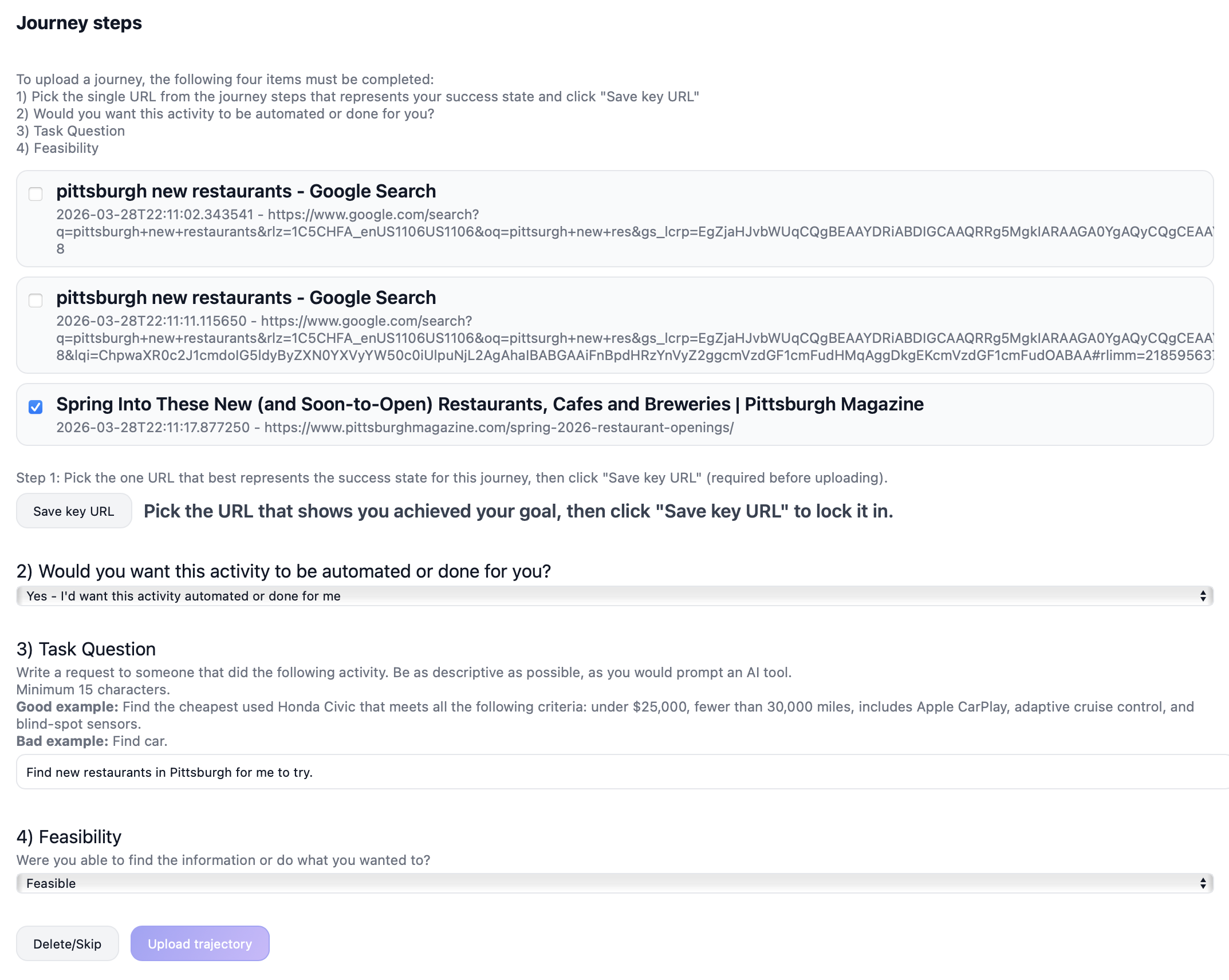}
    \caption{The annotation interface used by participants to label their Chrome browsing journeys. The interface guides participants through four steps: selecting the key URL representing the success state, indicating an automation preference, writing a descriptive task label, and judging feasibility.}
    \label{fig:apollo}
\end{figure}

Figure~\ref{fig:apollo} shows the desktop application used by participants to annotate their Chrome browsing journeys (Sec.~\ref{sec:task-composition}). For each journey, the interface displays the segmented URLs and guides the participant through four annotation steps: (1)~selecting the key URL that represents the task's success state, (2)~indicating an automation preference, (3)~writing a task label as they would prompt an AI tool, and (4)~judging feasibility.

\clearpage
\subsection{Journey Refinement} \label{app:refinement}

The 2,380 raw journeys collected from participants required substantial refinement before they could serve as building blocks for \odysseys tasks.

\paragraph{LLM screening} An LLM screens each journey along four dimensions: (1)~label accuracy (whether the task description matches what the visited URLs actually support), (2)~feasibility (whether the task can be completed without login credentials or additional personal context), (3)~login requirement (whether any step requires authentication), and (4)~overall quality. For each journey, the LLM also generates a refined label scoped to what the URLs support, along with notes justifying each assessment. This process judges 71.8\% of the original labels as inaccurate relative to the URLs. This high rate reflects a combination of factors: participants sometimes described their \textit{intent} rather than what the URLs show (e.g., ``find a good laptop'' when the URLs only show a single product page), submitted labels for browsing sessions that spanned multiple unrelated goals, or wrote underspecified descriptions that could not be verified against the visited pages. The LLM-refined labels address these issues by anchoring the task description to the observable URL evidence.

\paragraph{Manual review} Each journey is then manually reviewed by the authors through the annotation interface shown in Appendix~\ref{app:annotation-ui}. Reviewers independently verify URL selections, assess feasibility by manually searching the web, and choose between the original participant label, the LLM-refined label, or a custom label written by the reviewer. After filtering for feasibility, removing tasks that require login-gated flows, and discarding low quality tasks, 696 usable journeys (29.2\% of the original 2,380) remained.

\clearpage
\subsection{Task Composition Pipeline} \label{app:composition-pipeline}

The 696 refined journeys are short, single-site subtasks. To compose them into long-horizon \odysseys tasks, we first cluster related journeys, then use an LLM to chain subsets into coherent multi-step workflows.

\paragraph{Clustering} We embed each journey label using \texttt{text-embedding-3-small} \citep{openai_text_embedding_3_small}, reduce the embeddings to 15 dimensions with UMAP, and cluster them using HDBSCAN. UMAP preserves both local and global embedding structure, and HDBSCAN automatically determines the number of clusters, accommodates variable-density groups, and assigns outlier journeys to a noise class rather than forcing them into ill-fitting clusters. We then construct a \textit{theme graph} by connecting clusters whose centroid cosine similarity lies within a fixed range, filtering out both weakly related and near-duplicate clusters.

\paragraph{Chaining} Multi-step \odysseys tasks are composed by traversing this theme graph. For each task, we select a seed cluster at random, run a 1--2 hop BFS to collect 3--6 related clusters, and draw 3--4 representative journeys per cluster to form 12--24 candidate journeys. GPT-5.4 then selects and orders a subset into a coherent workflow, generating a natural-language prompt, step plan with transitions, rubric with verification procedures, and a coherence score in a single call.

\paragraph{Composition prompt} The system prompt below defines the task format and requirements (few-shot style examples are omitted for brevity):

\begin{quote}
\scriptsize
\begin{verbatim}
You are an expert at designing realistic, long-horizon web agent benchmark tasks.

You will receive a MENU of atomic web browsing tasks. Each has an ID, website,
description, cluster theme, and optional location. Your job is to compose a
SUBSET into a single coherent multi-step task that a real person would actually do.

CRITICAL REQUIREMENTS:
1. SEQUENTIAL FLOW: Each step must logically lead to the next. Later steps must
   USE or BUILD ON information from earlier steps.
2. UNIFIED GOAL: All steps serve one overarching purpose.
3. GEOGRAPHIC CONSISTENCY: If steps involve physical locations, they must be in
   the same city/region.
4. CROSS-SITE: Use at least 2 different websites.
5. INFORMATION DEPENDENCIES: At least 30% of steps should depend on a prior
   step's output.
6. NATURAL VOICE: The agent_prompt must sound like a real person talking to an
   assistant -- conversational, with personal context.

STYLE EXAMPLES: [8 few-shot examples spanning easy -> very_hard]

OUTPUT FORMAT: { goal, task_name, agent_prompt, steps, rubric, dependencies,
                 skills, primary_skill, deliverable, self_score, reasoning }
- Select exactly {target_length} steps for the task.
- You may SKIP menu items that don't fit.
- If you cannot form a coherent task, return {"steps": [], "self_score": 0}.
\end{verbatim}
\end{quote}

\noindent The user prompt provides the target difficulty, step count, and a menu of candidate journeys drawn from the related clusters:

\begin{quote}
\scriptsize
\begin{verbatim}
Target: 6 steps, hard difficulty

Available tasks:
  [A0] [google.com] Search for round-trip flights from Pittsburgh to LAX
    cluster: flight search and booking
  [A1] [booking.com] Find hotel near airport with overnight stay option
    cluster: hotel comparison and booking
  [A2] [maps.google.com] Check drive time between two locations
    cluster: route planning and navigation
  [A3] [enterprise.com] Search for car rental options at airport
    cluster: car rental comparison
  ...

Compose a coherent 6-step sequential workflow.
\end{verbatim}
\end{quote}

\clearpage
\subsection{Hard Long-Horizon CUA Task Pipeline}
  \label{app:hard-longhorizon-cua-pipeline}

  The refined journeys are mostly short, single-site tasks.
  To convert them into long-horizon tasks, we use each
  journey as an inspiration seed. The goal is to preserve the original user intent and data distribution
  while expanding the task into a realistic multi-hour browser
  workflow.

  \paragraph{Generation} We run GPT-5.4 for 2,380 refined journeys.
  For each journey, the model either rewrites the seed into one viable
  hard long-horizon CUA task or marks it as not viable. The rewrite is
  instructed to preserve the user goal, but is not tied the exact website
  or URL path. For example, a one-page lookup about whether Marriott
  has Taiwan rentals can become a broader task about planning and comparing stays across all hotel options for a future
  Taiwan trip. A
  single NYC bagel shop price lookup can become a comparison of similar popular bagel shops. If a seed cannot be expanded
  without artificial difficulty, it is rejected rather than inflated..

  \paragraph{LLM Judging} Each candidate rewrite is passed through a GPT-
  5.4 judge. Candidates that are artificial, too short, too vague, stale, or rubric-misaligned are rejected. Rejected candidates receive one
  retry when the source appears salvageable. The full run produced 567
  viable hard candidates from 2,380 refined journeys.

  \paragraph{Human QA} We visualize the generated candidates, sort
  them by quality and judge score, and manually review the strongest
  candidates. During QA, we remove tasks that feel artificially
  inflated, edit prompts that need more natural wording or clearer
  deliverables, and shortlist the tasks that best match the desired
  hard long-horizon CUA distribution. This process produced an 80-task
  final shortlist.

  \paragraph{Generation prompt} The system prompt below defines the
  rewrite requirements (in-context examples are omitted for
  brevity):

  \begin{quote}
  \scriptsize
  \begin{verbatim}
  Rewrite a browsing seed into one strong hard long-horizon CUA task.

  Core rule:

  - Treat the source journey as inspiration, not a spec.
  - Preserve the broader user goal and decision axis, not the literal
    site, URL,
    or page flow.
  - Source URLs and site metadata are hints, not rails.
  - Natural cross-site expansion is good when it better serves the
    goal on public
    pages.

  A strong rewrite:

  - reads like one believable first-person request
  - creates a whole browsing session: discover, compare, verify,
    synthesize
  - is plausibly multi-hour because the work itself is genuinely deep
  - uses explicit counts and measurable constraints
  - is browser-native: open or keep key tabs, compare menus/listings/
    maps/photos,
    and leave useful final pages open
  - uses public pages only and allows "not shown" for missing fields
  - for tiny lookup seeds, broadens only one level up into the real
    decision the
    user likely cares about, then stops
  - only asks for a document, sheet, or formal artifact when that
    genuinely helps
    the user goal; do not default to CryptPad or spreadsheets

  Avoid:

  - generic market-research inflation disconnected from the seed's
    real goal
  - awkward loyalty to the original site when a broader public-web
    task would be
    better
  - unrelated side quests added only to make it look hard
  - benchmark-y or robotic wording
  - turning a bounded lookup into a fake hard task just by adding more
    sources,
    ranking mechanics, evidence tabs, or a decision memo

  Time policy:

  - use the current date only internally to avoid stale or already-
    passed tasks
  - for travel, lodging, event, booking, or schedule tasks, prefer
    future-relative
    wording such as "for a future trip", "a few weeks from now", or
    "next season"
  - mark not viable if no credible public-page hard task exists
    without obvious
    distortion

  Return JSON only with these keys:
  { viable_hard, viability_reason, rewrite_rationale, flags,
  temporal_flags,
  confirmed_task, rubrics }

  Rubric rules:

  - viable hard rewrites must have 5 to 7 rubric items
  - rubric weights must sum to exactly 1.0
  - rubrics must verify both substantive output and browser evidence
  - every rubric must be directly traceable to something explicitly
    asked in the
    user-facing prompt
  - mirror prompt quantities exactly
  - do not add hidden deliverables, hidden sites, hidden tab counts,
    hidden fields,
    or hidden evidence requirements that the prompt never asked for
    \end{verbatim}
    \end{quote}

  \noindent The user prompt provides the seed journey metadata and
  curated hard in-context examples:

  \begin{quote}
  \scriptsize
  \begin{verbatim}
  Source journey:

  - Original label: Use Apollo Bagels' menu to calculate the total
    cost of ...
  - Canonical refined label: ...
  - Source site: ...
  - Source website field: ...
  - Source reference length: ...
  - Source URLs:
      - ...

  Curated hard exemplars:
  EXEMPLAR 1
  Task ID: ...
  Website: ...
  Summary: ...
  Why strong: ...
  Rubric headlines:

  - ...

  Use the source for its goal and decision axis, not its literal flow.
  If the seed is too atomic, broaden naturally into a more 
  useful same-axis browsing session, 
  including cross-site work when that better serves the goal.

  If the seed is a tiny lookup, prefer the broader decision 
  the user probably cares about unless the narrow 
  local frame is clearly the natural need. 
  Usually this means broadening one 
  practical level up and no further.

  For atomic seeds, default to a recommendation, shortlist, or
  decision memo with
  kept-open evidence tabs. Only ask for a document or spreadsheet if
  the task
  genuinely needs one because the scope is large enough.

  Keep time-sensitive tasks durable and future-relative when
  appropriate. Rubrics may organize grading, but 
  they must not add hidden work beyond what the prompt asked for.
  \end{verbatim}
  \end{quote}

  \clearpage
\subsection{Odysseys Task QA Interface} \label{app:task-qa}

\begin{figure}[h]
    \centering
    \includegraphics[width=\linewidth]{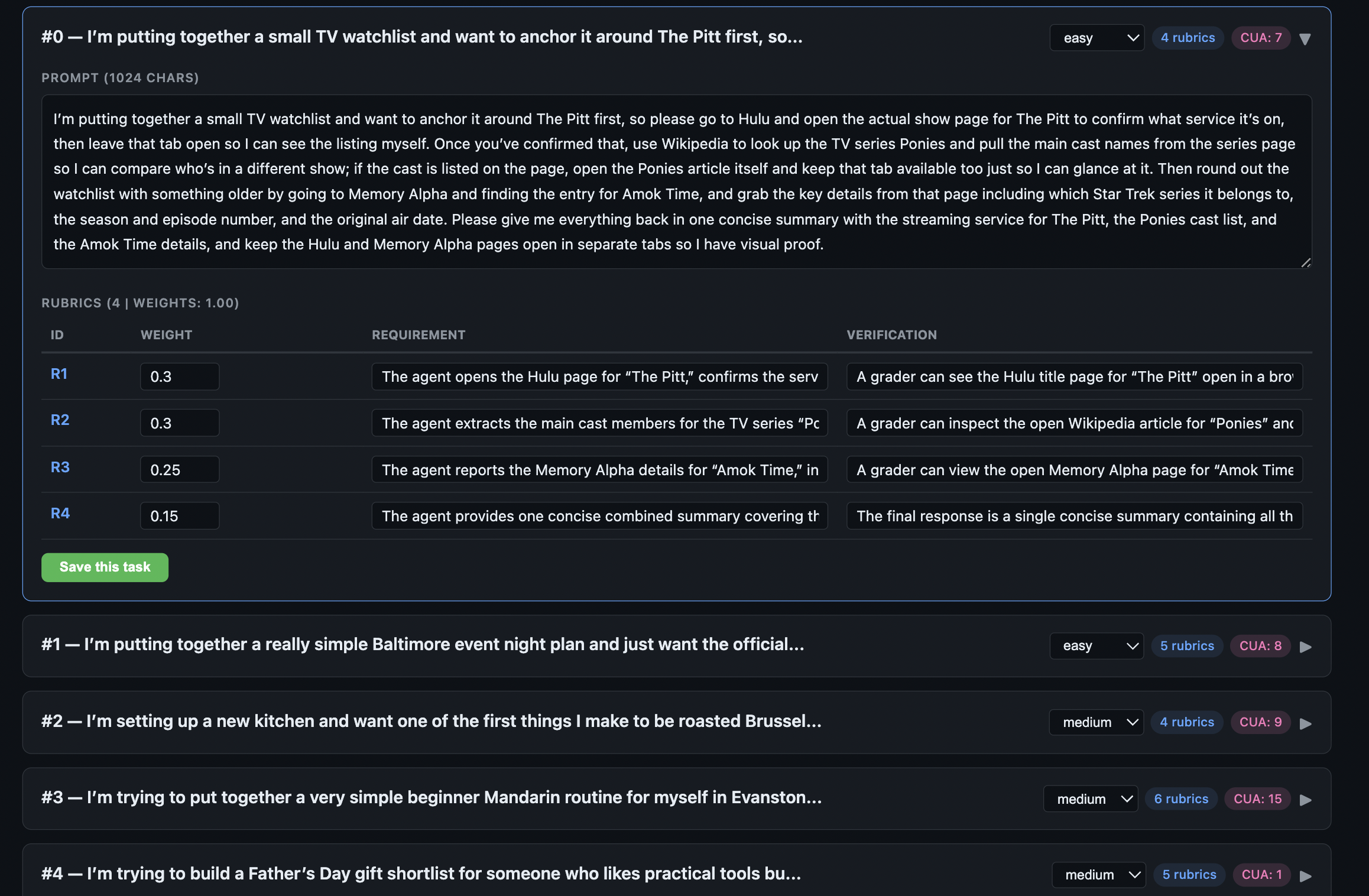}
    \caption{The \odysseys QA interface used for manual review of chained tasks. Reviewers can expand each task to inspect the full prompt and individual rubric items with weights, requirements, and verification criteria.}
    \label{fig:odysseys-qa}
\end{figure}

After composition and prompt rewriting (Sec.~\ref{sec:rubrics}), every chained \odysseys task is reviewed by the authors using the QA interface shown in Fig.~\ref{fig:odysseys-qa}. The interface lists all tasks with their difficulty level, rubric count, and score. Expanding a task reveals the full prompt and individual rubric items with weights, requirements, and verification criteria, allowing reviewers to verify coherence and actionability before the task enters the benchmark.

\clearpage

\subsection{Additional Evaluation Details}

For all of the models that we ran, we adopted the OSWorld\footnote{https://github.com/xlang-ai/OSWorld} runner using Chrome as the default app, starting at the google.com URL. The models are prompted with the initial task description to complete the task, and the implementation varies depending on the official OSWorld implementation for that model. For the Qwen3.5 models, as they frequently do not output a final summary of the task and any information requested, we additionally prompted it at the end to produce a summary with the prompt in Table~\ref{tab:qwen-summary-prompt}.

\subsection{Rubric Judge Prompt} \label{app:rubric-judge-prompt}

We score each trajectory with one judgment call per rubric item. Every call gives the judge the user task (for context only), a single rubric item with its requirement and verification description, the agent's full action history, and the trajectory screenshots in chronological order. The judge responds with a short rationale and a binary \texttt{success}/\texttt{failure} verdict, which we map to a rubric score of 1 or 0.

\paragraph{Judge prompt} The system prompt below fixes the evaluation rules and the expected response format:

\begin{quote}
\scriptsize
\begin{verbatim}
You are an expert evaluator of web-navigation agent trajectories.

You will receive:
- The user task (for context).
- ONE specific rubric item with a requirement and a verification description.
- The agent's full action history (one line per step).
- Every screenshot from the trajectory, in chronological order.

Your goal is to decide whether this single rubric item is satisfied by the
trajectory.

Evaluation rules:
- Judge ONLY the one rubric item you are given; ignore all other implicit
  requirements.
- Ground your judgment in what the screenshots and actions actually show.
  Do not invent state.
- Filtering / sorting / form requirements must be *applied and confirmed*
  (visible effect in results) to count as satisfied.
- If the agent was blocked (captcha, access denied, etc.) and therefore
  could not satisfy the rubric, report failure.

Respond in exactly this format:

Thoughts: <your reasoning, citing specific steps/screenshots>
Status: "success" or "failure"
\end{verbatim}
\end{quote}

\noindent The user message fills in the current task, the rubric item under evaluation, and the per-step action history, with the trajectory screenshots attached as image parts after the text:

\begin{quote}
\scriptsize
\begin{verbatim}
User Task (context only): {task}

Evaluate ONLY this rubric item:
Rubric ID: {rubric_id}
Requirement: {requirement}
Verification: {verification}
Weight: {weight}

Full Action History:
{action_history}

Screenshots attached below: {num_screenshots} (trajectory had {total_steps}
total step(s)).

Decide whether the rubric ({rubric_id}) is satisfied. Use the required
'Thoughts:' / 'Status:' format.
\end{verbatim}
\end{quote}

\subsection{200-Step Scaling on Hard Tasks} \label{app:200step-scaling}

\begin{figure}[h]
    \centering
    \includegraphics[width=\linewidth]{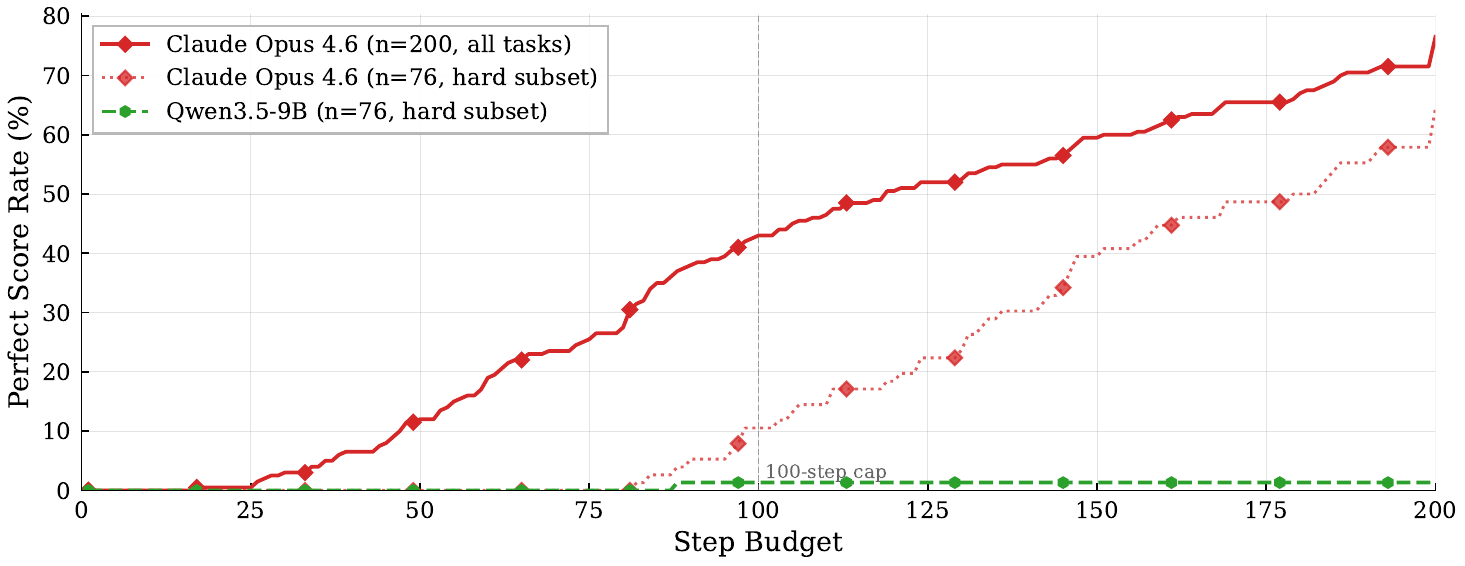}
    \vspace{-0.1in}
    \caption{Perfect rubric rate vs.\ step budget at a 200-step budget. Claude Opus 4.6 is shown on the full 200-task set (120 authored/composed + 80 LLM-generated hard) and continues to improve well past the original 100-step cap (vertical dashed line), climbing from 44.5\% to 76.5\%; 14.5\% of Opus runs still exhaust the 200-step budget, suggesting further headroom. For an apples-to-apples comparison, we also show Opus and Qwen-3.5-9B restricted to the 76-task hard subset for which Qwen has 200-step runs: Opus climbs to 64.5\%, while Qwen flatlines at 1.3\%, indicating that additional test-time compute alone is not enough to close the gap on hard long-horizon tasks.}
    \label{fig:success-vs-steps-200}
\end{figure}

\paragraph{Increasing max steps to 200} To probe whether the 100-step cap is the binding constraint on the eventual success of frontier models, we ran Opus 4.6 with a 200-step budget. The perfect task rate increases from \textbf{44.5\% in a 100-step budget to 76.5\% in 200 steps} (86/200 $\to$ 153/200), and the fraction of runs that exceed the step limit in the 200 step run is 14.5\% (29/200). Fig.~\ref{fig:success-vs-steps-200} (appendix) shows that Opus's curve continues to rise monotonically across the full range (44.5\% at 100, 50.5\% at 120, 59.5\% at 150, 67.0\% at 180, 76.5\% at 200), meaning that additional step budget translates into meaningful gains for capable agents on genuinely long-horizon tasks. 
To upper-bound the reachable performance, we optimistically assume every run that hit the 200-step cap would eventually reach perfect given unbounded steps. Even under this assumption, Opus 4.6 tops out at 86.0\% on the 200-task set, and at least $\sim$14\% of tasks fail for reasons other than insufficient test-time compute. For 200 steps, Opus 4.6 gets a trajectory efficiency score of 1.08\%, which is very similar to its score with 100 steps (1.06\%), indicating that its performance is not deteriorating past 100 steps, and it is still doing useful work.

\paragraph{Failure modes that persist at 200 steps}
We found three qualitative failure modes account for the majority of residual errors in Opus 4.6's performance.
\textit{(1) Information gathering without deliverable production.} The largest category involves the agent thoroughly browsing and collecting data but never compiling the required summary, spreadsheet, or CryptPad document before terminating. For example, on the New York birthday-weekend task, Opus 4.6 spent 140 steps comparing Arlo hotel rates across every May weekend and filtered Yankees home games on Ticketmaster, but never opened individual event pages to record the required price/venue fields and never produced the requested final trip-planning summary, exhausting all 200 steps.
\textit{(2) Abandoning progress for an incorrect programmatic shortcut.} On research-heavy tasks, Opus 4.6 occasionally abandons UI interaction mid-trajectory in favor of scripting a terminal-based solution that then fails to recover. In a clinical trial task, after opening 7--8 ClinicalTrials.gov tabs, the agent switched to fetching trials via the ClinicalTrials.gov API from a Python shell; the subsequent document was never populated and all 7 rubrics received zero. \textit{(3) High-fanout tasks that run out of breadth} Tasks requiring parallel coverage of many items tend to stall after completing a subset. For the academic job-market sweep , Opus 4.6 compiled the details of the Indiana University graduate program in $\sim$175 steps but ran out of budget before reaching the required cross-program comparison summary. These failures indicate that current agents struggle with the structure of tasks and would likely benefit from subagents or explicit step-aware planning.

\paragraph{Qwen 3.5 flatlines even at 200 steps}
Qwen 3.5-9B exhibits a qualitatively different curve, as extending the budget from 100 to 200 steps yields essentially zero additional progress on the 80 hard tasks. Within the same 200-step run used for Opus above, Qwen 3.5-9B (with a summarization prompt; Appendix~\ref{tab:qwen-summary-prompt}) reaches 1/76 perfect tasks by step 100 and does not solve a single additional task between steps 100 and 200. On the identical task set, Opus gained +55 pp (10\% $\to$ 65\%) over the same 100-step extension. Only 10.5\% of Qwen runs actually hit the 200-step cap, and the optimistic upper bound assuming that all capped runs eventually succeeded is just 11.8\%, meaning that additional computation alone cannot close the gap. Qualitatively, Qwen failures differ from Opus's. Rather than running out of budget on a coherent plan, Qwen frequently thrashes between tabs without satisfying any coverage criterion, indicating a capability limit rather than a step-budget limit.

\begin{table}[t]
\centering
\begin{tabular}{p{0.95\linewidth}}
\hline
\textbf{Qwen Summary Prompt} \\
\hline
\begin{minipage}{\linewidth}
\vspace{0.5em}
\ttfamily
You are a helpful assistant. You previously used a computer to complete a task. Now answer the user's follow-up question using only the information you gathered during the task.

\vspace{0.5em}

Respond with plain text only. Do NOT output any <tool\_call> blocks, function calls, or action descriptions.

\vspace{0.5em}

The current date is \today.
\vspace{0.5em}
\end{minipage} \\
\hline
\end{tabular}
\caption{Prompt used to elicit a final summary from Qwen3.5 models}
\label{tab:qwen-summary-prompt}
\end{table}

\subsection{Full Task Descriptions for Table~\ref{tab:odyssey-examples}} \label{app:examples}

The following are the complete, unabridged task prompts for the six representative tasks shown in Table~\ref{tab:odyssey-examples}, listed in order of difficulty. These are the exact instructions given to the computer-use agent.

\paragraph{Easy: Kitchen setup (Brussels sprouts, Le Creuset, dishwasher air gap)}
\begin{quote}
\small
I'm setting up a new kitchen and want one of the first things I make to be roasted Brussels sprouts, so could you start on Google and find me a recipe that clearly uses both Parmesan and balsamic vinegar, then open the actual recipe page and note the title, oven temperature, and cook time because I want to make sure the cookware I buy fits that kind of roasting setup. Once you've got that recipe open, head to Le Creuset and look for a light green Dutch oven, and specifically check whether the 5.5 qt size is offered in that color so I can see if it would work for recipes in that range; please open the product page itself and leave it open so I can look at the color and size options on the page. While you're at it, I'm also sorting out kitchen appliances before I start cooking, so go to YouTube and find a practical video about whether a dishwasher installation needs an air gap, open the video page, start playing it, and tell me the main decision points like when an air gap is required, when a high loop is used instead, and why the air gap exists in the first place. Please keep the recipe tab, the Le Creuset product tab, and the YouTube video tab open in separate tabs so I can compare everything visually afterward.
\end{quote}

\paragraph{Easy: Weather risk snapshot (Baltimore, Syracuse, Rittman, Mount Holly)}
\begin{quote}
\small
I'm trying to decide whether driving this week is a bad idea, so can you build me a quick weather risk snapshot that starts with what it feels like right now and then widens out to the bigger trouble spots? First, on Google, search for Baltimore, Maryland weather and grab the current temperature plus the plain-English condition like cloudy, sunny, rain, or whatever it says, just so I have a baseline for home conditions. Then go to Wunderground and look up Syracuse, New York, and check the 10-day/7-day style forecast to find the lowest temperature expected over the next 7 days, including which day it happens, because I want to compare that colder destination against Baltimore. After that, use the National Weather Service forecast page for the Rittman, Ohio area near Marshallville and tell me what the current forecast says and whether there are any active alerts posted there, since that would really affect an Ohio leg of the drive; please open the actual forecast page and leave it visible so I can see the alert area and forecast text myself. Finally, go to the NWS Mount Holly page, find the winter forecast graphic, and report the snowfall amount shown there so I can tell whether the Mid-Atlantic part looks like a nuisance event or something more serious; if the graphic opens separately, leave that tab open too so I can look at the map. In the end, send me a short location-by-location summary with the key weather risk for Baltimore, Syracuse, Rittman/Marshallville, and the Mount Holly region.
\end{quote}

\paragraph{Medium: Japan trip planning for a Canadian in the U.S.}
\begin{quote}
\small
I'm a Canadian citizen living in Pittsburgh, PA, and my passport expires in about 3 months, so I'm trying to get everything sorted before a 2-week tourist trip to Japan. Could you start on the official Government of Canada site and find the passport renewal process for a Canadian living in the U.S., including the exact renewal form I'd need, the supporting documents, photo rules, whether I need a guarantor or references, the fee in CAD, how I'm supposed to submit it from the U.S., and the current processing time, because I need to know if this is realistic before I book anything. Once you have that, use the official Canadian embassy/consulate pages to figure out which Canadian mission is closest to Pittsburgh, Pennsylvania 15222 that handles passport services, and open the actual office page so I can see the address, passport service hours, and whether I need an appointment or have to use some booking request process; please leave that page open. After that, check Japan's official Ministry of Foreign Affairs site to confirm whether a Canadian passport holder going to Japan for tourism for 2 weeks needs a visa, and note any conditions or exceptions that matter. Then go to the Government of Canada travel advisory page for Japan and tell me the current advisory level plus any highlighted health, safety, or entry notes, and keep that advisory page open in another tab so I can look at it myself. Finally, compare travel insurance options on PolicyAdvisor.com and Kanetix.ca for this situation: a Canadian citizen currently living in the U.S. who wants coverage connected to travel to Japan, and I mainly want to see whether either site shows plans that would actually work for someone based in the U.S. rather than Canada, so please capture provider names, medical emergency coverage, trip cancellation/interruption if shown, and any residency or eligibility restrictions. If either site has useful quote or results pages, open the most relevant options in separate tabs so I can compare them visually. At the end, give me a concise summary that ties all of this together and clearly points out any uncertainty, especially around insurance eligibility for a Canadian living in the U.S.
\end{quote}

\paragraph{Medium: Tesla Model 3 lease evaluation in Los Angeles}
\begin{quote}
\small
I'm trying to sanity-check whether moving ahead with a Tesla Model 3 lease in Los Angeles is actually manageable month to month, so start on Google and look up current Tesla Model 3 lease pricing for the Los Angeles area, including the lease term, due-at-signing amount, and any discounts, tax credits, or rebates you can find, because I want a realistic baseline instead of just a headline number. Once you've got that monthly lease figure, use it as a reference point and go to the Los Angeles Craigslist site, specifically the San Gabriel Valley section, and find at least three trailer listings that look like plausible live-in fallback options under \$10,000; open the actual posting pages in separate tabs so I can see the photos and verify the listings are still live, then note each one's title, price, and location. After that, go to Zillow and look for an LA-area rental whose monthly price is in the same ballpark as the Tesla lease payment you found, so I can compare whether paying for housing at that level makes more sense than taking on the car; open the actual Zillow listing page and leave it open so I can check the photos and map myself. In the end, give me a short comparison that includes the Tesla lease deal, the three Craigslist trailer backups, and the Zillow rental option that's closest in monthly price to the lease.
\end{quote}

\paragraph{Hard: 30 MLB stadiums summer road trip}
\begin{quote}
\small
I'm daydreaming about doing a ridiculous-but-fun summer baseball trip where I see exactly one game at all 30 MLB stadiums, and I want you to build the whole thing in a way I could actually use. Start on MLB.com and pull the official summer schedule so we can choose one real game date at each stadium, and please lean toward matchups where I might get to see stars I care about most like Shohei Ohtani, Aaron Judge, and Ronald Acuna Jr.\ whenever that's realistically possible. As you're picking games, open the actual game or team schedule pages in separate tabs for a few representative stops so there's visible proof the dates are live, and keep the key schedule tabs open so I can glance at them later. Once you've got the 30 stadium/date choices, use Google Flights to figure out the smartest sequence between stops and compare flights versus driving for each leg, using whatever is cheaper and more practical in summer, because I want this to feel like a real budget-conscious trip instead of fantasy routing. After that, use Booking.com to find one hotel option for each game night that's reasonably close to the stadium---something like within about 2 miles if possible and not outrageously priced for a solo traveler---and open at least a couple of the actual hotel listing pages with photos/maps so I can visually sanity-check the neighborhoods. Then use Yelp to find at least one must-try local food spot near each stadium, ideally something iconic to that city or ballpark area, and open a few of the restaurant pages so I can see that they're real places with reviews. Finally, put everything into a CryptPad Document in one organized itinerary with each stadium listed exactly once, the chosen game and matchup, whether it includes Ohtani, Judge, Acuna, or another notable player, the travel leg before it with the cheaper mode and estimated cost, one hotel with estimated nightly price, one food pick, and running totals so I can see what this insane summer would actually cost. Leave the finished CryptPad Document open at the end, and if you create any comparison tabs along the way, keep the most useful ones open so I can review them.
\end{quote}

\paragraph{Hard: Banana bread recipe synthesis}
\begin{quote}
\small
I want to develop the best banana bread recipe. Look up the top 10 recipes online (by engagement, popularity, reviews) and compare the recipes (e.g.\ composition of ingredients, additions, cooking method), identifying and highlighting similarities and unique points that make each recipe good. Keep the most unique or highly reviewed 3 recipes in open tabs so I can reference them, and make sure at least one has a YouTube video (also keep this video open and start playing it). Then, from these three, create the best recipe you can combining aspects of these and provide me with step by step instructions.
\end{quote}

\clearpage
\subsection{Human Agreement Annotation Interface} \label{app:human-agreement-ui}

Figure~\ref{fig:human-agreement-ui} shows the trajectory viewer used by annotators to evaluate rubric satisfaction for each agent run (Sec.~\ref{sec:eval}). The interface displays the agent's step-by-step actions alongside screenshots, reasoning, and task rubrics. For each rubric item, the annotator marks whether it was satisfied during the trajectory. Ambiguous cases are flagged for discussion.

\begin{figure}[h]
    \centering
    \includegraphics[width=\linewidth]{figures/human_agreement_UI.png}
    \caption{The trajectory viewer used for human agreement annotation. Annotators step through the agent's actions (left panel), view screenshots and action details (center), and judge each rubric item as pass or fail (right panel).}
    \label{fig:human-agreement-ui}
\end{figure}

\clearpage
\subsection{Further Qualitative Results}

The following figures illustrate the surprising agent behaviors discussed in Sec.~\ref{sec:analysis}.

\begin{figure}
    \centering
    \includegraphics[width=\linewidth]{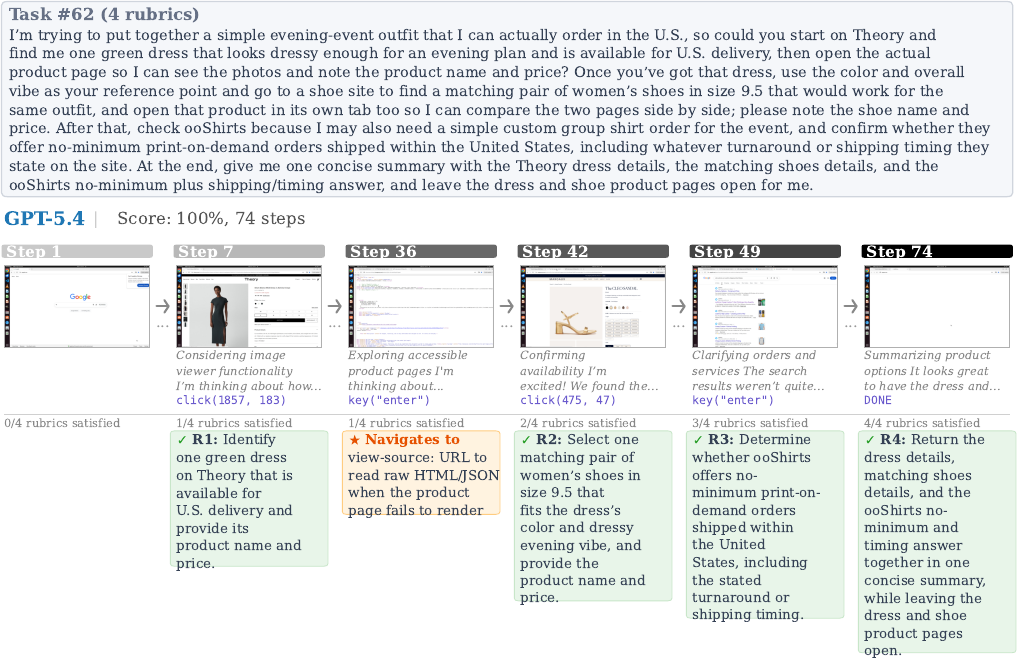}
    \caption{When \texttt{margauxny.com} rendered blank due to JavaScript failures, the GPT-5.4 agent typed \texttt{view-source:https://margauxny.com/products/...} directly in the address bar, then used \texttt{ctrl+f} to search for \texttt{variants}, \texttt{InStock}, \texttt{40.5}, and \texttt{US 9.5} in the raw HTML. From the embedded JSON-LD schema, it decoded EU 40.0 = US 9.5, confirmed that the SKU was in stock, and reconstructed the direct variant URL, all without the product page ever visually rendering.}
    \label{fig:gpt-62-viewsource}
\end{figure}

\begin{figure}
    \centering
    \includegraphics[width=\linewidth]{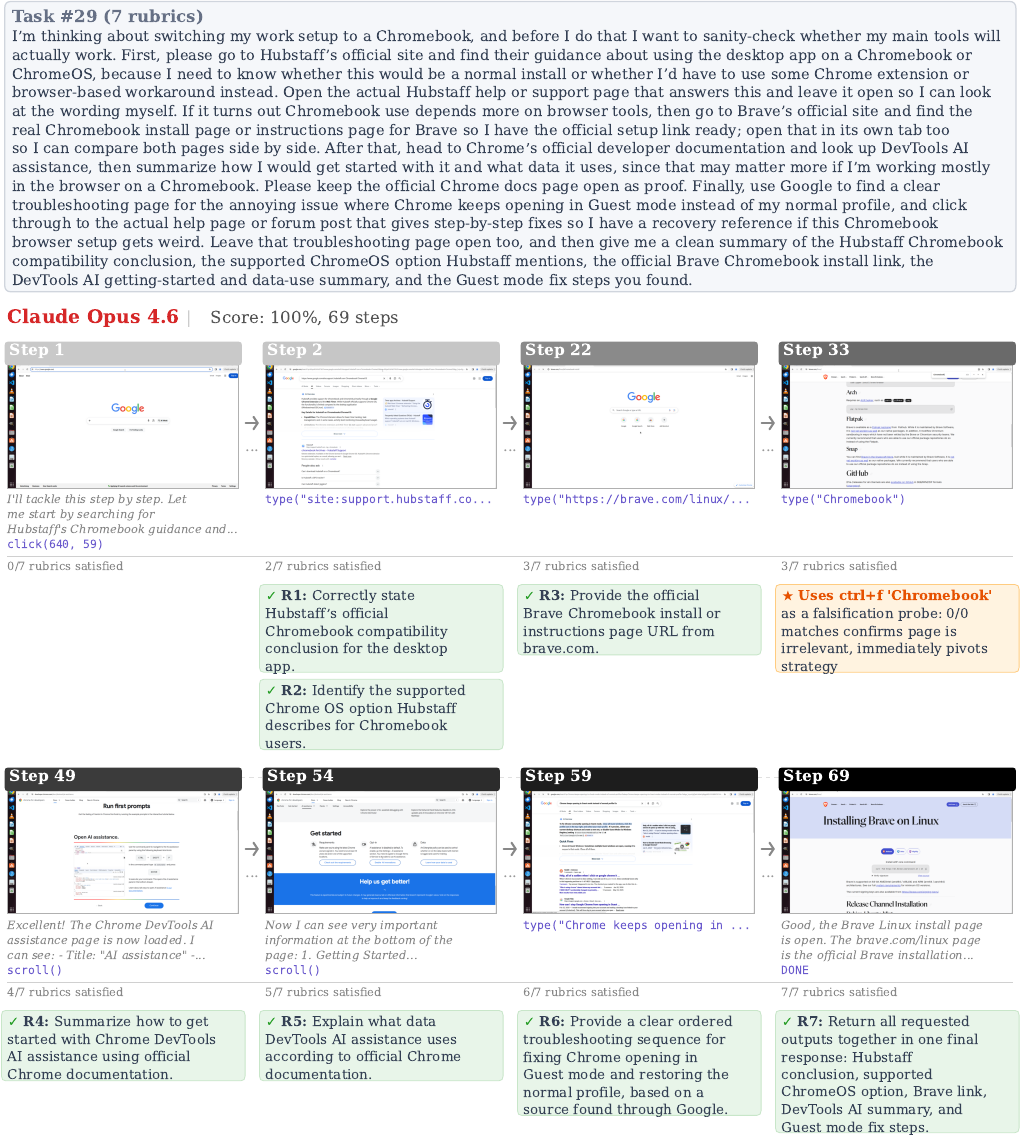}
    \caption{After navigating to \texttt{brave.com/linux}, Opus used \texttt{ctrl+f} to search for \texttt{Chromebook} and observed \texttt{0/0} matches, confirming that the page did not cover the topic at all, and immediately pivoted to a different strategy. Rather than scrolling to verify, it treated the absence of a match as decisive evidence that the page was unhelpful.}
    \label{fig:claude-29_ctrlf_probe}
\end{figure}

\begin{figure}
    \centering
    \includegraphics[width=\linewidth]{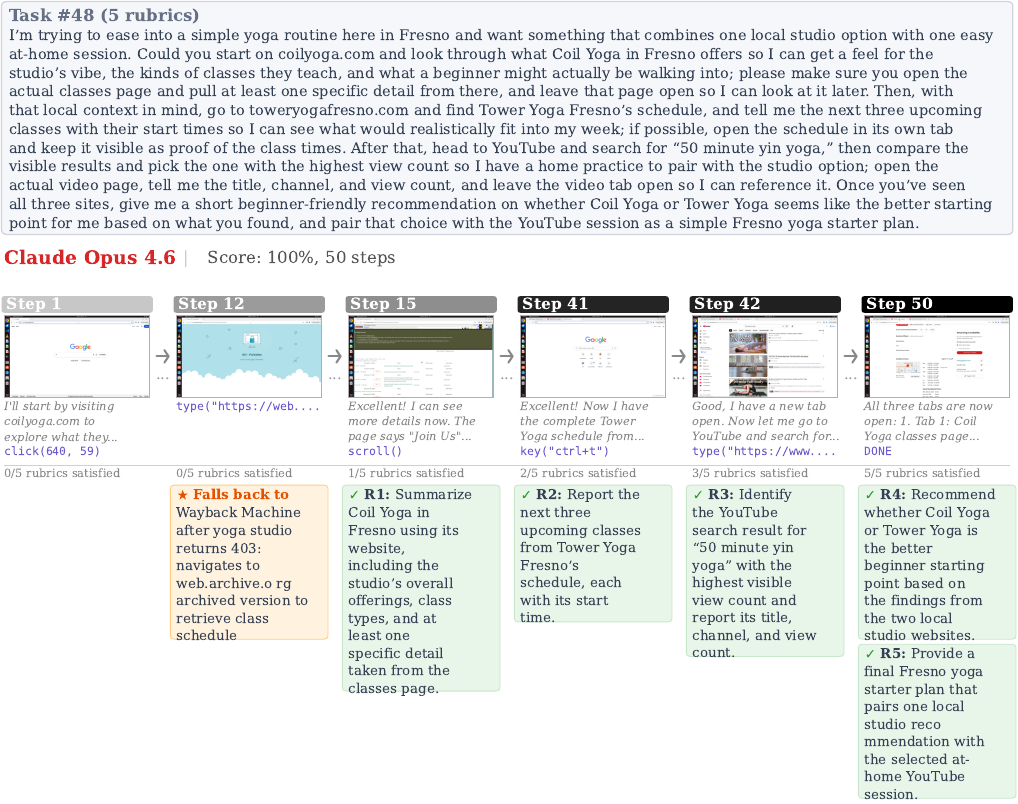}
    \caption{Both yoga studio websites returned \texttt{403 Forbidden}. Opus immediately navigated to \texttt{https://web.archive.org/web/2024/https://coilyoga.com/classes/} and repeated the same strategy for Tower Yoga, successfully retrieving archived pages from June 2024 that contained the full class schedule information.}
    \label{fig:claude-48_wayback}
\end{figure}

\end{document}